\documentclass{sig-alternate-05-2015}
\usepackage{amsfonts}
\usepackage[colorlinks=true, citecolor=blue, urlcolor=cyan, linkcolor=red]{hyperref}
\usepackage{amsmath,amssymb,amscd,epsfig,amsfonts,rotating}
\usepackage{graphicx}
\usepackage{epstopdf}
\usepackage{subfigure}
\usepackage{multirow}
\usepackage{booktabs}
\usepackage{xcolor}
\usepackage{url}
\usepackage{amsmath,amscd,amsbsy,amssymb,latexsym,url,bm}
\usepackage{epsfig,epstopdf,graphicx,subfigure}
\usepackage{enumitem,balance}
\usepackage{wrapfig}
\usepackage{mathrsfs, euscript}
\usepackage{algorithm}
\usepackage{algorithmic}
\usepackage{amssymb}
\usepackage{ifpdf}
\usepackage{makecell}

\newcommand{\bs}{\boldsymbol}

\usepackage{color}

\begin{document}

\setcopyright{rightsretained}

%

\conferenceinfo{Neu-IR '16 SIGIR Workshop on Neural Information Retrieval,}{July 21, 2016, Pisa, Italy}


%
\conferenceinfo{Neu-IR '16 SIGIR Workshop on Neural Information Retrieval,}{July 21, 2016, Pisa, Italy}

\title{Learning text representation using recurrent convolutional neural network with highway layers}

%
%
%
%
%

\numberofauthors{1} 
%
\author{
%
%
\alignauthor
Ying Wen$^{1,2}$, Weinan Zhang$^{1}$, Rui Luo$^{1}$, and Jun Wang$^{1,2}$ \\
       \affaddr{$^{1}$University College London, $^{2}$MediaGamma Limited} \\
       \affaddr{London, UK}\\
       \email{\{ying.wen, w.zhang, r.luo, j.wang\}@cs.ucl.ac.uk}
}

\maketitle
\begin{abstract}
Recently, the rapid development of word embedding and neural networks has brought
new inspiration to various NLP and IR tasks. In this paper, we describe a staged
hybrid model combining Recurrent Convolutional Neural
Networks (RCNN) with highway layers. The highway network module is incorporated in the
middle takes the output of the bidirectional Recurrent Neural Network (Bi-RNN) module in the first stage
and provides the Convolutional Neural Network (CNN) module in the last stage with
the input. The experiment shows that our model outperforms common neural network
models (CNN, RNN, Bi-RNN) on a sentiment analysis task. Besides, the analysis of
how sequence length influences the RCNN with highway layers shows that our model
could learn good representation for the long text.
\end{abstract}

\printccsdesc

\section{Introduction}
Learning good representations for text, such as words, sentences and documents, is essential for
information retrieval (IR) and natural language processing (NLP) tasks like text
classification and sentiment analysis, which has attracted
considerable attention from both academic and industrial communities \cite{bengio2013representation}.

It is common to use bag-of-words or bag-of-ngram to represent text \cite{croft2010search} and train
the models based on such representation. However, each word or n-gram is a
unique feature and their interactions and the whole word order in the text are not preserved in the text representation,
which limits the functionality of the learning models based on such representation.

In recent years, the word embedding and neural network models have brought new
solutions to learn better representations for NLP and IR tasks. The simplest
one is Bag-of-Word-Vectors (BOW vector) model. But Landauer et al.
\cite{landauer2002computational} estimates that 20\% of the meaning of a text
comes from the word order. Therefore, these models are still oversimplified
because of the loss of order information. Neural language model \cite{bengio2006neural}
was then proposed to leverage word embedding representation to infer the next-word
distribution, but it still fails to fully utilize the sequence of the context words.

Recurrent Neural Networks (RNN) can take word order into account, but it suffers
from the problem that later words make more influence on the final text representation than
former words. However, for sentiment analysis tasks, which is the studied problem of this work, the important words indicating the
right sentiment may occur anywhere in the document.

Convolutional Neural Networks (CNN) are naturally capable of solving this problem.
CNN treats each words fairly by using the max-pooling layer \cite{kalchbrenner2014convolutional}.
Besides, comparing
with RNN, which is more natural to process sequence information and get fixed
length output, CNN uses sliding windows with different width and filters
to perform the feature mapping, then pooling is used to get fixed length output. In
addition, CNN is comparatively simple, efficient and has achieved strong empirical
performance on the text classification jobs \cite{kim2014convolutional}.
But CNN also has problems, such as determining the window width and too many filter
parameters.

To combine the advantages from Recurrent Neural Network and Convolutional Neural
Network,  Lai et al. 2015 proposed Recurrent Convolutional Neural Network (RCNN)
\cite{lai2015recurrent}. This model applies bi-directional recurrent from
the beginning to the end of the document to capture the contexts around each word.
Then, combine a word and its context to present a word, where concatenate the
around contexts and word embedding and filter will be used to calculate the
latent semantic vectors. Finally, max-pooling is used to capture the most
important factor and make fixed length sentence representation.

In this work, we propose a Recurrent Convolutional Neural Network with Highway
Network (RCNN-HW), which incorporates Highway Networks with
Recurrent Convolutional Neural Networks. In RCNN-HW model, one highway layer is
introduced as intermediate layer between bidirectional RNN and the CNN, which
helps to select features individually for each word representation.
Therefore, our model further optimises the original RCNN model and
achieves better performance.

To summarize, the contributions of our work as follows:
\begin{itemize}
\item We first propose the new architecture that incorporates Highway Networks with
Recurrent Convolutional Neural Networks.
\item We apply the proposed model on the sentiment analysis task and achieved
superior performance over strong baselines.
\item We investigate how sequence length influences the performance of
different neural network models, and find RCNN and RCNN-HW models have better
performance on longer text.
\end{itemize}

\begin{figure}[t]
\epsfig{file=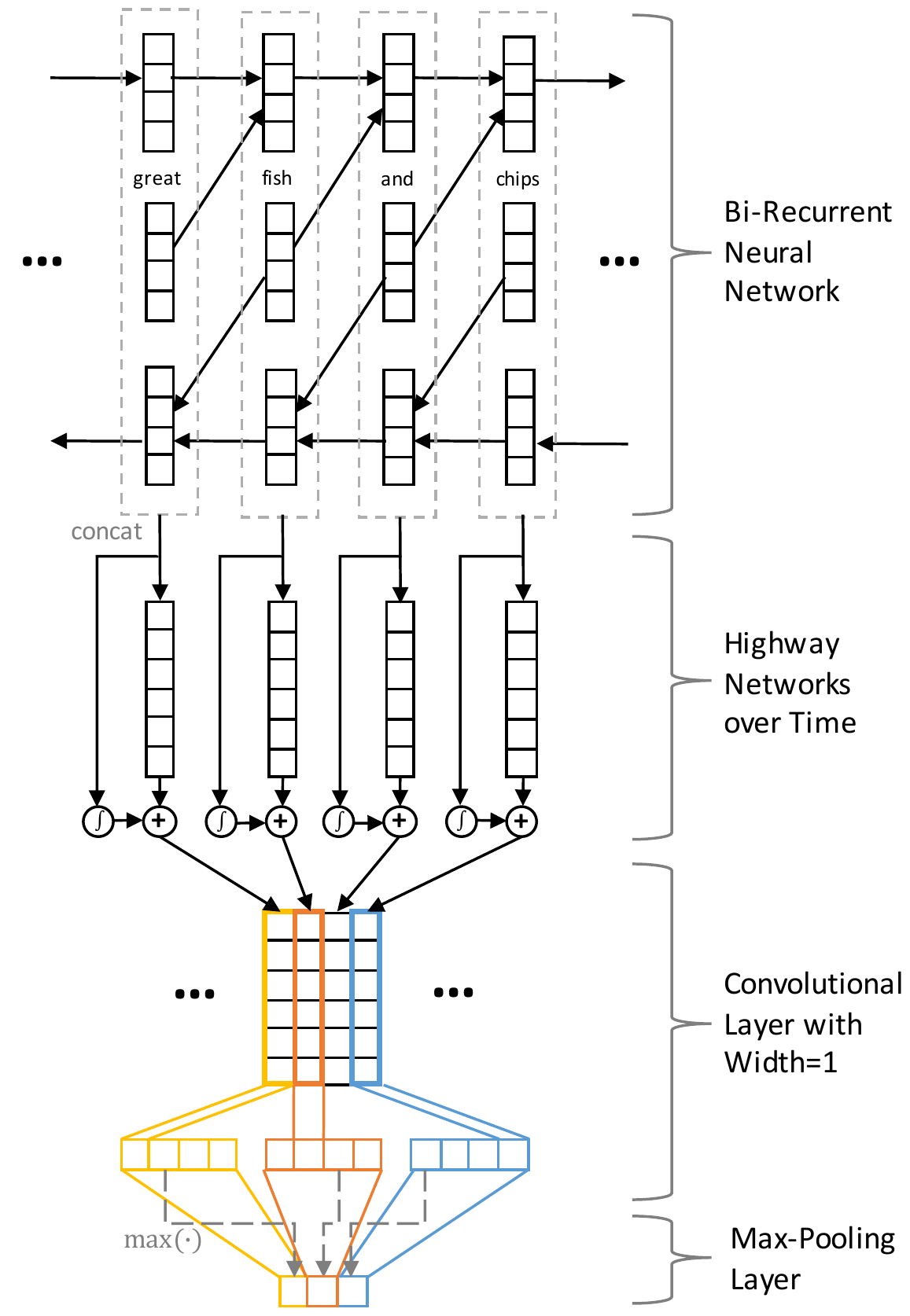, width=3.33in}
\caption{Architecture of Recurrent Convolutional Neural Network with Highway Layers.}
\label{rcnn_highway}
\end{figure}

\section{Model}

In this section, we propose a staged hybrid model combining RCNN with highway
layers, which is illustrated in Figure \ref{rcnn_highway}. As is shown, the
highway network module in the middle takes the output of the bidirectional RNN
module in the first stage and provides the CNN module in the last stage with
the input.


\subsection{Recurrent Neural Networks}

It is quite intuitive and straightforward to adopt RNN for learning sequential
data due to its nature: RNN is able to process current data instances while
taking into account preserved historical information as if it has memory.
Among those approaches for enhancement of the long-term memory,
Long Short-Term Memory (LSTM) \cite{hochreiter1997long} and its variants are of particular
interest due to the great success in practice. Thus we use as the building
block of our network a newly proposed variant, Gated Recurrent Unit
(GRU) \cite{cho2014learning}, for its simplicity and competitive performance.
The expressions of GRU is given as follows:
\begin{equation}
\begin{split}
\boldsymbol{r}_t &= \sigma(\mathbf{W}_r\boldsymbol{x}_t + \mathbf{U}_r\boldsymbol{h}_{t-1} + \boldsymbol{b}_r) \\
\boldsymbol{z}_t &= \sigma(\mathbf{W}_z\boldsymbol{x}_t + \mathbf{U}_z\boldsymbol{h}_{t-1} + \boldsymbol{b}_z) \\
\boldsymbol{\tilde{h}}_t &= \mathrm{tanh}(\mathbf{W}_h\boldsymbol{x}_t + \mathbf{U}_h(\boldsymbol{r}_t\odot\boldsymbol{h}_{t-1}) + \boldsymbol{b}_h) \\
\boldsymbol{h}_t &= \boldsymbol{z}_t\odot\boldsymbol{h}_{t-1} + (1-\boldsymbol{z}_t)\odot\boldsymbol{\tilde{h}}_t
\end{split}
\end{equation}
where $\odot$ represents element-wise multiplication and $\mathbf{W},\mathbf{U},\boldsymbol{b}$ are input weights, recurrent weights and biases, respectively.

We exploit the struture of bidirectional RNN in order to extract both
forward-passing and backward-passing information: the context information from
the left and right hand side of a particular word. Hence a new context-containing
representation $\boldsymbol{\tilde{x}}_t$ of a word is obtained by simply
concatenating the output of the bidirectional RNN ($\boldsymbol{h}_t^{\leftarrow},
\boldsymbol{h}_t^{\rightarrow}$) with the word embedding $\boldsymbol{x}_t$:
\begin{equation}
\boldsymbol{\tilde{x}}_t=[\boldsymbol{h}_t^{\leftarrow}\|\boldsymbol{x}_t\|\boldsymbol{h}_t^{\rightarrow}]
\end{equation}

\subsection{Highway Networks}

A one-layer highway network is introduced as the intermediate layer between the
bidirectional RNN and the CNN to select features individually for each word
representation. The highway layer is expressed as follows:
\begin{equation}
\boldsymbol{y}_t = \boldsymbol\tau\odot g(\mathbf{W}_H\boldsymbol{\tilde{x}}_t + \boldsymbol{b}_H) + (1-\boldsymbol\tau)\odot\boldsymbol{\tilde{x}}_t
\end{equation}
where $g$ is nonlinear function and
$\boldsymbol\tau = \sigma(\mathbf{W}_\tau\boldsymbol{x}_t + \boldsymbol{b}_\tau)$
represents ``transform gate''. Note that the design of highway connection is
quite similar with that of GRU's ``update gate'' $\boldsymbol{z}$, which is
essentially a variant of ``leaky integration'' \cite{bengio2013advances}. It
allows part of the input information to be carried unchanged to the output
while the rest to go through some (nonlinear) transformations. The experiments
indicate that this kind of structure will help to extract substantial information.


\subsection{Convolutional Neural Networks}

In recent years, CNN has achieved great success in CV and has been proved to
be effective in various NLP and IR tasks. A CNN architecture is
generally composed by a stack of distinct layers mapping input to output
via some piece-wise differentiable function. In order to extract new features from the input (phrases/sentences in our case), a convolutional layer essentially applies
a set of learnable filters to the input, which have small receptive fields: they are in fact a set of masks with different window sizes $h$, i.e. they select $h$ consecutive words for the neurons in the layer. For instance, feature $\boldsymbol{c}_i$ is extracted from a window of
word representations with size $h$: $\boldsymbol{y}_{i:i+h-1}$ according to the following expression:
\begin{equation}
c_i = f(\boldsymbol{w}_{\text{conv}} \cdot \boldsymbol{y}_{i:i+h-1} + b_{\text{conv}})
\end{equation}
where $f$ is a nonlinear function which in our case is the ``Rectified Linear Unit'' (ReLU), $\boldsymbol{w}_{\text{conv}}, b_{\text{conv}}$ are the weight and bias.

The filter $f$ will be applied to the word representations of the whole text with length $n$ via a sliding window of size $h$ to establish the feature map:
\begin{equation}
\bs{c} = [c_1,c_2,...,c_{n-h+1}]
\end{equation}

Next to the convolution layer, a max-pooling layer is applied to the feature map, which extracts the most significant feature $\widehat{c} = \max\{\boldsymbol{c}\}$.

The proposed model utilises multiple filters with window size $h=1$ to extract features for establishing the representation of the whole text. Note that although it is a common practice to exploit different window sizes, we fix the size to $1$ since the bidirectional RNN and highway layer in the earlier layers have already obtained the context information around each word and thus we are free from
applying filters in different window sizes.


\section{Experiment}
To demonstrate the effectiveness of the proposed model, we perform
the experiment on IMDB dataset \cite{maas2011learning} which contains 25,000 movie
reviews for training. Each review is labelled by human with 1 represents positive sentiment
and 0 represents negative sentiment. The review in the dataset has
267.9 words in average and standard deviation of review length is 198.8. To test the performance of
learned text presenation of models, we run sentiment prediction on this dataset and report
the accuracy evaluation metric. Besides, we are particularly interested in the
relationship between input sequence length and performance of the model, and set up
the experiment to test the performance of the model over different input sequence lengths.

We compare our model RCNN with highway layers (RCNN-HW) with the following baseline
neural network models for document level sentiment prediction:

\begin{itemize}
\item \textbf{Sum-of-Word-Vector (COW)}: simply sums up all the word embeddings as text representation.
\item \textbf{LSTM}: takes the average of the LSTM's hidden states of all words is used as text representation \cite{hochreiter1997long}.
\item \textbf{Bi-LSTM}: similar to LSTM but exploits bidirectional LSTM \cite{schuster1997bidirectional} on the sequence to get text representation.
\item \textbf{CNN}: performs 1-demensional convolution followed by 1-demensional max-pooling
with multiple filters on the sequence \cite{kalchbrenner2014convolutional, kim2014convolutional}.
\item \textbf{CNN+LSTM}: combines CNN and RNN (LSTM) by using CNN's output as LSTM's input.
\item \textbf{RCNN}: uses bidirectional RNN's output as CNN's input \cite{lai2015recurrent}.
\end{itemize}

We use neural network models above to learn the text representation and adopt softmax output to
make prediction. For the neural network models, hyperparameter tuning has great influence on its
performance. For all the models we have done grid search with reasonable hyperparameters
and tried different stochastic gradient descent training methods (e.g. adam, RMSprop, adadelta) \cite{kingma2014adam}.
For our model, RCNN with highway layers\footnote{The code is avaliable in \url{https://github.com/wenying45/deep_learning_tutorial/tree/master/rcnn-hw}}, we choose `RMSprop' as training
method with batch size of 32, the hidden layer dimension of 32, and the filter number of 256.
Besides, we also tested the the performance of different neural network models with multiple
input sequence lengthes, the RCNN based models get better performance with longer
sequence length, in our experiment, we choose the best performance of the model
over different sequence lengthes. Note, we did not apply pre-training methods like
using pre-trained word2vec \cite{NIPS2013_5021}, and regularizer such as
Dropout \cite{srivastava2014dropout} which may bring further improvement \cite{kim2014convolutional}.

\begin{table}[]
\centering
\caption{Accuracy for different neural models on IMDB dataset}
\label{imdb_table}
\begin{tabular}{ll}
\cline{1-1}
\hline
\multicolumn{1}{c}{Model} & Accuracy \\ \hline
COW &0.890          \\
LSTM               &0.885          \\
Bi-LSTM            &0.881          \\
CNN+LSTM            &0.890          \\
CNN                &0.895          \\ \hline
RCNN               &0.900          \\
RCNN-HW  &\textbf{0.903}          \\ \hline
\end{tabular}
\end{table}

\section{Results and Discussion}

\begin{table}[]
\centering
\caption{Accuracy for RCNN models with/without highway layers}
\label{highway_table}
\begin{tabular}{ll}
\cline{1-1}
\hline
\multicolumn{1}{c}{RCNN based model} & Accuracy \\ \hline
Without Highway Layers  &0.900     \\
One Highway Layers &\textbf{0.903}          \\
Two Highway Layers &\textbf{0.903}          \\
One MLP Layer      &0.899          \\ \hline
\end{tabular}
\end{table}

The experiment results can be found in Table \ref{imdb_table}.
We compare our RCNN with highway layers with orginal RCNN model and find
that the performance with highway layers are always better than those without. To
quantitatively investigate the effect of highway network, we set up an
experiment with RCNN based models. We train the RCNN model without highway layers, with one
layer highway, two layers highway and one layer MLP respectively with the same parameters, then
compare their accuracy. The results are shown in Table \ref{highway_table}. We can find
that having one to two highway layers is important, but more highway layers do
not improve the performance. Besides, one MLP layer does not gain more
improvement than higway layers. We think the reason why the highway layer
works is because the highway layer helps select the features of word representation.

Besides, we also compare with CNN, RNN and COW models. Empirical experiments show that
CNN models usually yield better performance than RNNs which include LSTM and Bi-LSTM.
It may be caused by RNN based models, especially the LSTMs, are hard to be trained,
these models are really sensitive to the hyperparameters and latter words make
more influence on the final text representation than former words
in RNN models. CNN models usually perform remarkably well on many NLP and
IR tasks \cite{hu2014convolutional,kalchbrenner2014convolutional,kim2014convolutional,zhang2015sensitivity}.
CNN based models use a fixed window of words as contextual information and the performance of
a CNN is influenced by the window size. A small window may result in a loss of
some long-distance patterns, whereas large windows will lead to data sparsity
 \cite{chung2014empirical,lai2015recurrent}. However, RCNNs outperform the CNN and the
CNN-LSTM model, we know that convolution structure can capture local context information, and
recurrent can capture global information. We consider that in CNN-LSTM model, it does
not make sense to use convolution layer and max-pooling layer before recurrent
layer. Such architecture means using local information features extracted from CNN as RNN's
input, but local information features do not have sequence relationship. As for
why RCNN based models outperform CNN, we think our context-containing representation
of a word obtained by simply concatenating the output of the bidirectional RNN
with the word embedding which has encoded context information would be a
better input to the CNN comparing with original word embedding.

\begin{figure}[h]
\centering
\epsfig{file=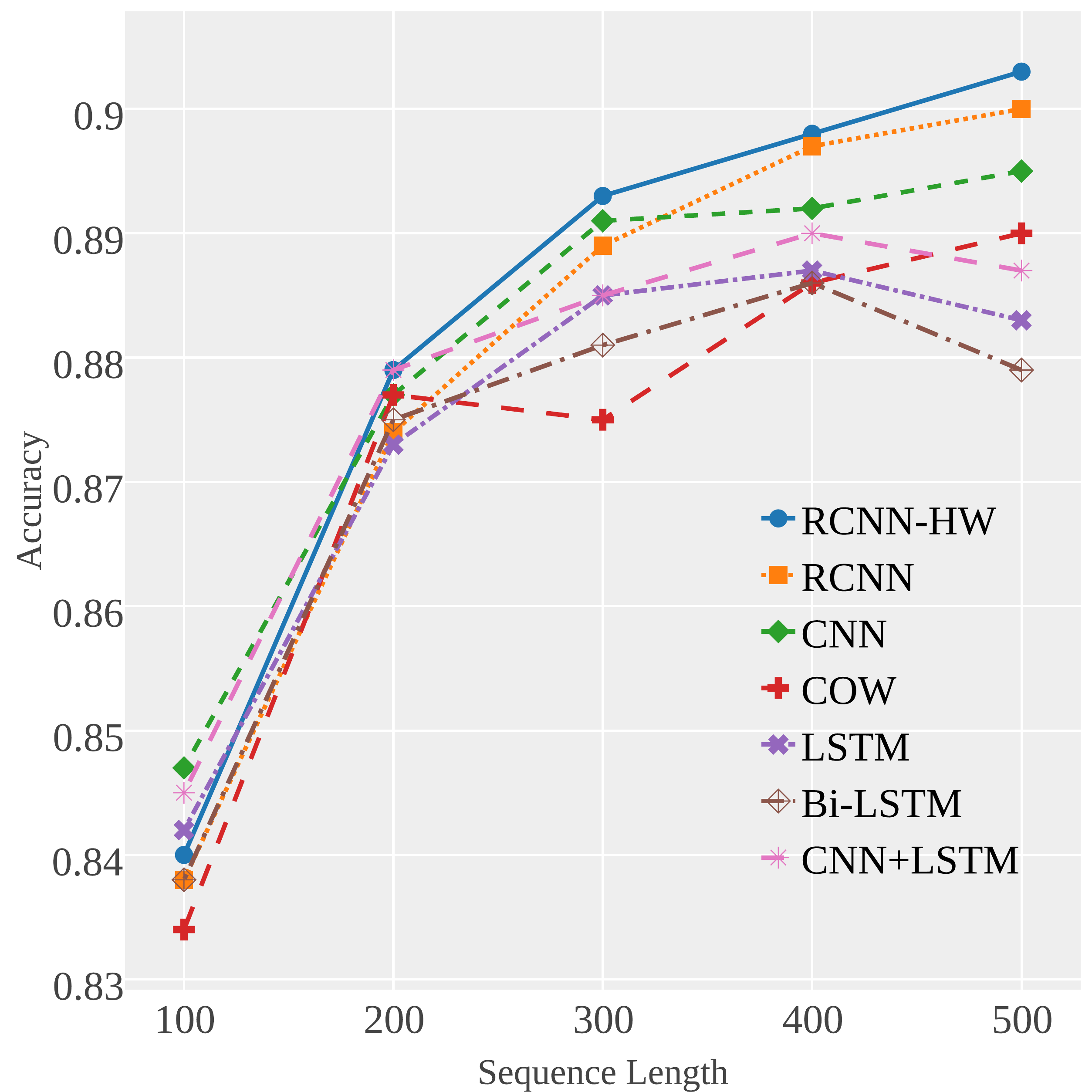,,width=0.8\columnwidth}
\caption{Accuracy curve for how sequence length influences the performance of different neural network models}\label{seq_len}
\end{figure}

We further investigate how the input sequence length influences the
performance of different neural network models. The average text length in our
experiment distaste is 267 and the standard deviation of the text length is around 200.
Therefore, we consider the input sequence length from 100 to 500 with step 100
to train the neural network models. The prediction accuracy for different neural
network models with various input sequence length are shown in Figure \ref{seq_len}.
In the figure, we can observe that when the sequence length is less than 200, all
models get bad performance since the short input length results in
information loss. As the input sequence length increases, models receive longer
input sequence which may bring more useful information as well as more noise
information. From the figure, we can find that RCNN based models,
especially our RCNN-HW model, obtain much higher improvement from longer input sequences.
Meanwhile, the other models get lower improvement and even get worse performance than
shorter input sequence length. We think this is because RCNN based models
can preserve longer contextual information thanks to their recurrent
structure and introduce less noise for their convolution structure with max-pooling.
And our RCNN-HW can further reduce the noise with a highway layer to perform
feature selection, which brings the best performance in longer input sequence lengthes.

We can take an example to illustrate why our RCNN-HW model outperformS than others. There
is a positive review ``I borrowed this movie despite its extremely low rating,
because I wanted to see how the crew manages to
animate the presence of multiple worlds. As a matter of fact, they didn't - at
least, so its seems ... But I closed my eyes. When I opened them again - he was gone.''
in the dataset,
and it's length is 498. This review can be easily classified as negavtive thus the
double negative and the long length. Simple model can hard to capture this type of
structure and handle the long-term denpendecy, so as COW, CNN, RNN models, they
predict this sample as a negative review. But our RCNN-HW model can deal with
this siuation and makes right predictbbbbion.

\section{Conclusion}
We have introduced a recurrent convolutional neural network with highway layers to
learn text representation for sentiment analysis. The experiment demonstrates
that our model RCNN-HW not only outperforms CNN and RNN models, but also works better than the
RCNN without highway layer. Besides, since our model yield better performance on
longer text, it would be interesting to see if the model introduced in this paper
works well in learning the text representation for the other NLP and IR tasks with relatively long
documents.

\section{Acknowledgments}
The authors would like to thank the anonymous reviewers for
their helpful feedback and suggestions.

%
\bibliographystyle{abbrv}
\bibliography{sigproc}  
\end{document}